\documentclass[sigconf]{acmart}

\renewcommand\footnotetextcopyrightpermission[1]{}
\def\ie{{\em i.e.}}
\def\eg{{\em e.g.}}
\def\etal{{\em et al.}}
\newcommand{\figref}[1]{Fig. \ref{#1}}
\newcommand{\tabref}[1]{Tab. \ref{#1}}
\newcommand{\equref}[1]{(\ref{#1})}
\newcommand{\secref}[1]{Section \ref{#1}}

\newcommand{\mc}[1]{\mathcal{#1}}

\newcommand{\br}[1]{\bm{\mathrm{#1}}}

\usepackage{epstopdf}

\usepackage{ragged2e}
\usepackage{booktabs}
\usepackage{color}
\usepackage{multirow}
\usepackage{bm}
\usepackage{bbm}
\usepackage[misc]{ifsym}

\usepackage{multirow}
\usepackage{float}

\usepackage{times}
\usepackage{amsmath}

\usepackage{amssymb}

\AtBeginDocument{%
  \providecommand\BibTeX{{%
    \normalfont B\kern-0.5em{\scshape i\kern-0.25em b}\kern-0.8em\TeX}}}

\setcopyright{acmcopyright}
\copyrightyear{2020}
\acmYear{2020}
\setcopyright{acmlicensed}
\acmConference[MM '20]{Proceedings of ACM Multimedia conference	(MM20)}{12 - 16 October 2020}{Seattle, United States}
\acmBooktitle{Proceedings of ACM Multimedia conference (MM'20)}
\acmPrice{15.00}
\acmDOI{10.475/123_4}
\acmISBN{978-1-4503-9999-9/18/06}

\begin{document}
\fancyhead{}
\title{Is Depth Really Necessary for Salient Object Detection?}


\author{Jiawei Zhao$^{1,\dagger}$, Yifan Zhao$^{1,\dagger}$, Jia Li$^{1,*}$, Xiaowu Chen$^{1}$}
\thanks{$^{\dagger}$Jiawei Zhao and Yifan Zhao contribute equally to this work.\\
$^{*}$Jia Li is the corresponding author (E-mail: \textsuperscript{}jiali@buaa.edn.cn). Url:\textsuperscript{}https://cvteam.net/}
\affiliation{%
	\institution{$^1$State Key Laboratory of Virtual Reality Technology and Systems, SCSE, Beihang University}
	}






\begin{abstract}
    
	Salient object detection (SOD) is a crucial and preliminary task for many computer vision applications, which have made progress with deep CNNs. Most of the existing methods mainly rely on the RGB information to distinguish the salient objects, which faces difficulties in some complex scenarios. To solve this, many recent RGBD-based networks are proposed by adopting the depth map as an independent input and fuse the features with RGB information. Taking the advantages of RGB and RGBD methods, we propose a novel depth-aware salient object detection framework, which has following superior designs: 1) It only takes the depth information as training data while only relies on RGB information in the testing phase. 2) It comprehensively optimizes SOD features with multi-level depth-aware regularizations. 3) The depth information also serves as error-weighted map to correct the segmentation process. 
	With these insightful designs combined, we make the first attempt in realizing an unified depth-aware framework with only RGB information as input for inference, which not only surpasses the state-of-the-art performance on five public RGB SOD benchmarks, but also surpasses the RGBD-based methods on five benchmarks by a large margin, while adopting less information and implementation light-weighted. The code and model will be publicly available.
\end{abstract}



\maketitle

\section{Introduction}

Salient object detection (SOD) aims to detect and segment objects that attract human attention most visually. With the proposals of large datasets~\cite{ju2014depth,peng2014rgbd,niu2012leveraging,yan2013hierarchical,li2015visual,wang2017learning} and deep learning techniques~\cite{he2016deep,long2015fully}, recent works have made significant progress in accurately segmenting salient objects,
which can serve as an important prerequisite for a wide range of computer vision tasks, such as semantic segmentation \cite{lai2016saliency}, visual tracking \cite{hong2015online}, and image retrieval \cite{shao2006specific}.

\begin{figure}[t]
	\begin{center}
		\includegraphics[width= \linewidth]{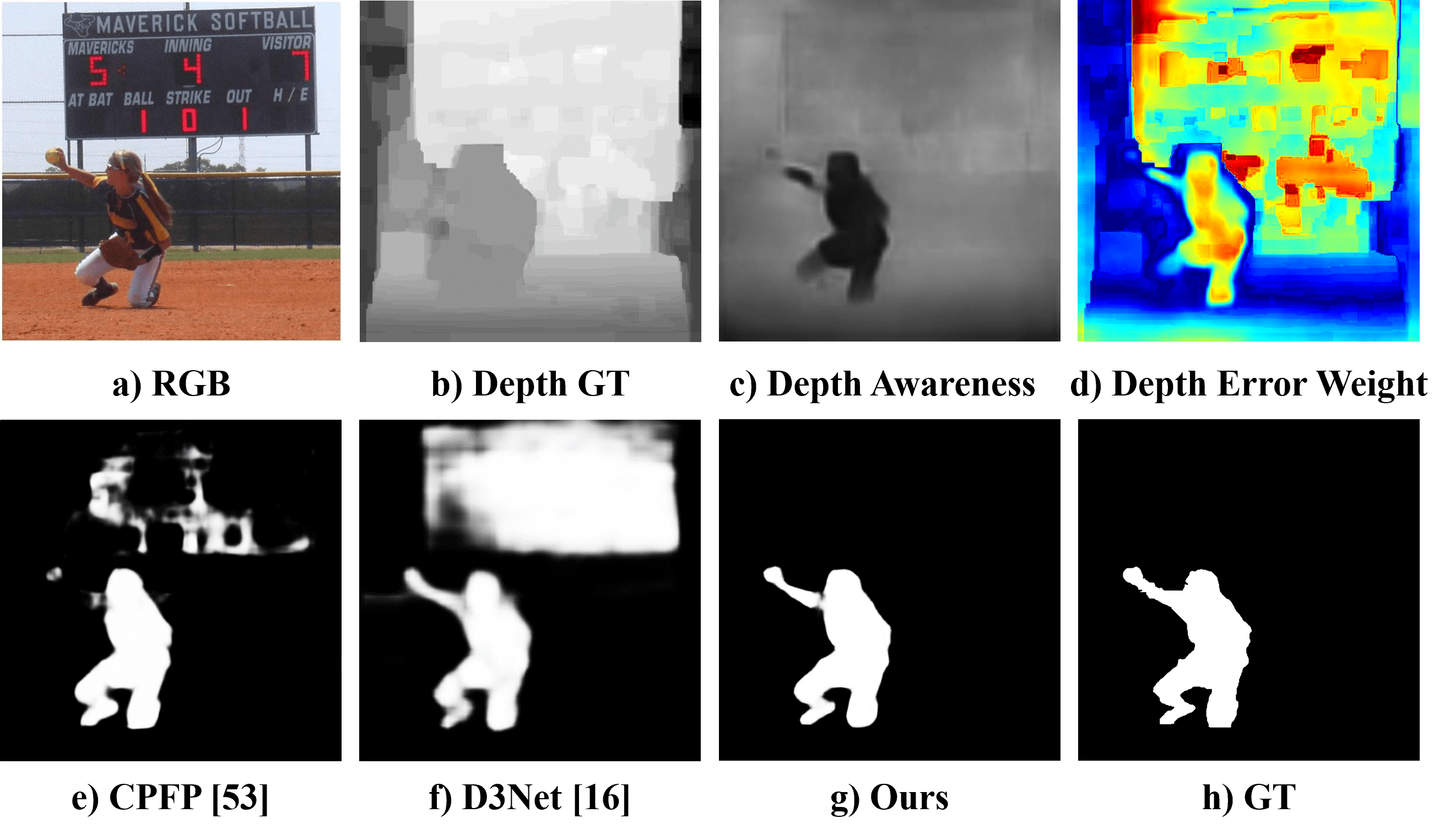}
		\caption{Motivation of our depth-aware salient object detection. b): captured depth groundtruth. c): predicted depth awareness by DASNet. d): depth-aware error weights for salient correction. e) and f) are generated by two RGBD SOD models.
		}\label{fig:motivation}
	\end{center}
\end{figure}

Recent years have witnessed significant progress in the field of salient object detection. Previous works~\cite{cheng2014global,yan2013hierarchical,klein2011center,liu2018picanet,F3Net,wu2019cascaded,zhao2019pyramid,su2019selectivity,qin2019basnet} take only the RGB information as inputs, which is relatively lightweight and can be easily trained end-to-end. For example, Wu \etal ~\cite{wu2019cascaded} propose a coarse-to-fine feature aggregation framework to generate saliency maps.
However, the reasoning of salient regions can not be well solved when there exist multiple contrasting region proposals or ambiguous object contours. Therefore, the depth information can be a complementary guidance to deduct the overlapping and viewpoint issues, which can be beneficial to salient object detection.

Combing the RGB information with the auxiliary depth inputs, recent research efforts~\cite{han2017cnns,piao2019depth,zhao2019contrast} have verified its effectiveness in improving the object segmentation process. 
These methods usually introduce an additional depth stream to encode depth map and then fuse the RGB stream and depth stream to deduct the salient objects. For example, Han \etal ~\cite{han2017cnns} propose a two-stream network to extract RGB features and depth features, and then fuse them with a combination layer. Piao \etal~\cite{piao2019depth} propose a two-stream network and fuse paired multi-level side-out features to refine the final saliency results.
The main drawbacks of RGBD-based methods are twofold. On the one hand, the additional depth branch introduces heavy computation costs compared to the methods with bare RGB inputs. On the other hand, the object segmentation process heavily relies on the acquisition of depth maps, which are usually unavailable in some extreme occasions or realistic industrial applications.
Keeping these cues in our mind, a natural concern arises: is depth information really necessary for salient object detection and what roles should depth play in salient object detection? 

Taking the essence and discarding the dregs of RGB and RGBD methods, we set out to create a unified framework, which only takes the depth information as supervision in the training stage. Hence the network can take only the RGB images as inputs, and meanwhile is aware of depth prior knowledge with the learnt network parameters. That is to say, we make use of depth information to regularize the learning process of salient object detection (See~\figref{fig:motivation}). First, we force the feature maps in different levels of network to be aware of depth information. This can be conducted in a multi-task learning trend when learning the object segmentation and estimating the depth map simultaneously. The estimated depth awareness map can be found in~\figref{fig:motivation} c). Although the estimated depth map is not highly accurate as captured one (in~\figref{fig:motivation} b)), but focuses on more contrastive depth regions, which are desirable for the segmentation process.
Second, the estimated depth awareness can also be considered as an indicator to find the most ambiguous regions. We calculate the logarithmic error map of the estimation and ground truth depth to generate an adaptive weight map in~\figref{fig:motivation} d). The network is further forced to pay more attention to pixels with higher error-weighted responses, hence some semantic confusions can be improved. Comparing to other state-of-the-art RGBD-based models~\cite{zhao2019contrast,fan2019D3Net} in~\figref{fig:motivation} e) and f), the proposed approach can better tackle the salient confusions while generating clear object boundaries. 

\begin{figure*}
	\begin{center}
		\includegraphics[width=1\textwidth]{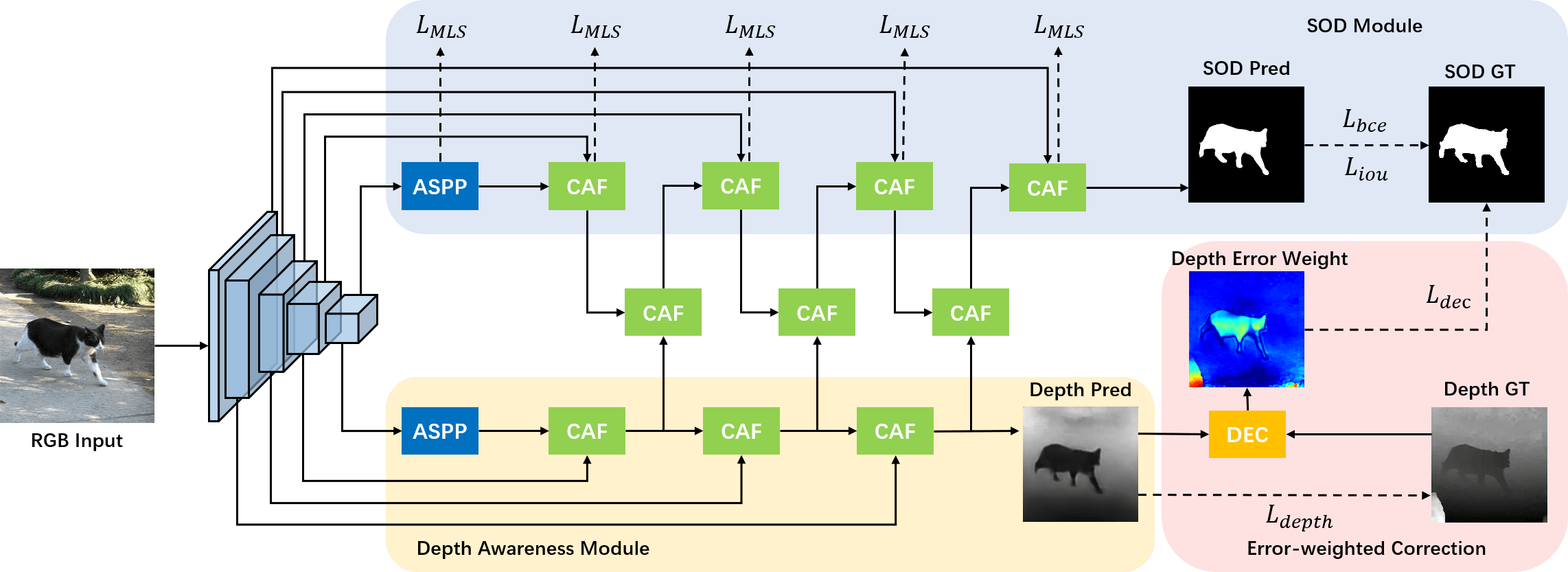}
		\caption{The overall architecture of our model. Our  depth-awareness SOD framework is mainly composed of three parts,~\ie, a salient object detection module, a depth awareness module and an error-weighted correction. ASPP denotes atrous spatial pyramid pooling. CAF denotes the proposed channel-aware fusion module. DEC denotes the proposed depth error-weighted correction. The dashed line denotes supervision. 
		}\label{fig:architecture}
	\end{center}
	
\end{figure*}

In this paper, we make three insightful designs to construct our framework in~\figref{fig:architecture}, which make full use of training data from multiple sources.~\ie, data from RGB source and RGBD source can be separately fed into this framework with different learning constraints to promote the final performance.
To achieve this framework, we first propose a depth awareness module, to regularize the features in different levels of the network stage while learning the object segmentation in the meantime. This forces the segmentation features to be aware of constrastive object in the depth of field. 
Second, we propose a generalized channel-aware fusion model (CAF) to aggregate the features from top to bottom levels in these two relevant branches. Then the final depth features and segmentation features are fused with the same CAF module in this coarse-to-fine scheme. Last but not least, we utilize a depth error-weighted map to emphasize the saliency ambiguous regions,~\ie, objects salient in images but not in depth, or vice versa. These regions are attached with more attention in the overall learning procedure to alleviate the object confusions and generate clear object boundaries. Experimental evidences demonstrate the effectiveness in promoting RGBD salient object detection with only RGB inputs and the potential in promoting RGB tasks with auxiliary training depth.

Contributions of this paper are summarized as follows: 1) We first set out a novel setting to use depth data as training priors to facilitate the salient object detection and propose a unified framework to solve this important problem. 2) We propose a channel-aware fusion model (CAF) to comprehensively fuse multi-level features, which can retain rich details and pay more attention to the significant features. 3) We propose a novel joint depth awareness module to facilitate the understanding of saliency and design a depth-aware error loss to mine ambiguous pixels. 4) Experimental evidences demonstrate that the proposed model achieves the state-of-the-art performance both on five RGBD benchmarks and five RGB benchmarks.

\section{RELATED WORK}

\textbf{RGB-based Salient object detection.}
Early traditional RGB SOD methods mainly rely on hand-crafted cues such as color constrant~\cite{cheng2014global}, texture~\cite{yan2013hierarchical} and local/global contrast~\cite{klein2011center}. Borji~\etal~\cite{borji2015salient} comprehensively review these methods for details with both deep learning and conventional techniques.
Recently CNN-based RGB SOD methods have achieved impressive improvements over traditional methods~\cite{cheng2014global,yan2013hierarchical,klein2011center}. Most of them follow an end-to-end architecture as shown in \figref{fig:introduction} a). Liu \etal{} \cite{liu2018picanet} utilize pixel-wise contextual attention to selectively attend to global and local context information. Wu \etal{} \cite{wu2019cascaded} propose a coarse to fine aggregation framework, which discards low-level features to reduce the complexity. Zhao \etal{} \cite{zhao2019pyramid} propose a pyramid feature attention network, which adopts channel-wise attention and spatial attention to focus more on valuable features. Su \etal \cite{su2019selectivity} propose a boundary-aware network to fuse the boundary and interior features with a compensation mechanism and an adaptive manner. Qin \etal \cite{qin2019basnet} design a hybrid loss to focus on the boundary quality of salient objects. Wei \etal \cite{F3Net} propose a cross-feature module to fuse features of different levels.

\textbf{RGBD-based Salient object detection.}
Although existing RGB methods have achieved very high performance, they might fail when dealing with complex scenarios, \eg, low contrast, occlusions. It is shown that depth is an important and effective cue for saliency detection~\cite{desingh2013depth} especially in these complex scenarios.
Existing RGB-D SOD methods mainly rely on extracting salient features from RGB image and depth map respectively, and then fuse them in the early or late network stages. Peng \etal{} \cite{peng2014rgbd} directly concatenate RGB-D pairs as 4-channel inputs to predict saliency maps. Han \etal{} \cite{han2017cnns} propose a two-stream network to extract RGB features and depth features, and then fuse them with a combination layer. Chen \etal{} \cite{chen2018progressively} propose a progressive fusion strategy in a coarse-to-fine manner. Zhao \etal{} \cite{zhao2019pyramid} propose a fluid pyramid integration strategy to make full use of depth enhanced features. Piao \etal{} \cite{piao2019depth} develop a two-stream network and fuse paired multi-level side-out features to refine the final salient object detection.

\textbf{Single image depth estimation.}
Monocular depth estimation can be divided into three categories according to the input: monocular video~\cite{wang2018learning,wang2019recurrent}, stereo image pairs~\cite{garg2016unsupervised,tosi2019learning} and single image~\cite{eigen2014depth,eigen2015predicting,laina2016deeper,fu2018deep,yin2019enforcing}, in which taking single image as input is the hardest case because there is no geometric information in only a single image. Thanks to the powerful deep networks like VGG~\cite{simonyan2014very} and ResNet~\cite{he2016deep}, single image depth estimation has been boosted to a new accuracy level. 
Eigen \etal \cite{eigen2014depth,eigen2015predicting} propose the first CNN-based framework for single image depth estimation, which applies a stage-wisely multi-scale network to refine depth estimation. Laina \etal{} \cite{laina2016deeper} introduce a fully convolutional architecture and design reverse Huber loss to smoothness effect of L2 norm. Fu \etal{} \cite{fu2018deep}  propose a spacing-increasing discretization strategy to discretize depth and recast depth estimation as an ordinal regression problem. Yin \etal \cite{yin2019enforcing} propose a global geometric constraint to improve the depth estimation accuracy.
As an important cue in many vision tasks, there are many works utilize multi-task learning to joint depth estimation and other per-pixel vision tasks, such as semantic segmentation~\cite{mousavian2016joint}, surface normal~\cite{yin2019enforcing}.

\section{METHODOLOGY}

\subsection{Overview}

\textbf{Depth-Awareness SOD Network.} 
In this section, we present a novel joint Depth-Awareness SOD Network (DASNet) for RGBD-based and RGB-based salient object detection tasks, which is mainly composed of three modules,~\ie, the SOD module, the depth awareness module and the depth error-weighted correction (see \figref{fig:architecture}). The first two modules share similar structures but focus on different tasks, which are supervised by saliency maps and depth maps respectively. The SOD module and the depth awareness module utilize our proposed channel-aware fusion model (CAF) to fuse high-level and low-level features. Taking these two branches into combination, we finally refine the saliency results by the proposed depth error-weighted constraint, which could mine hard pixels with the supervision of depth maps.

\begin{figure}
	\begin{center}
		\includegraphics[width= \linewidth]{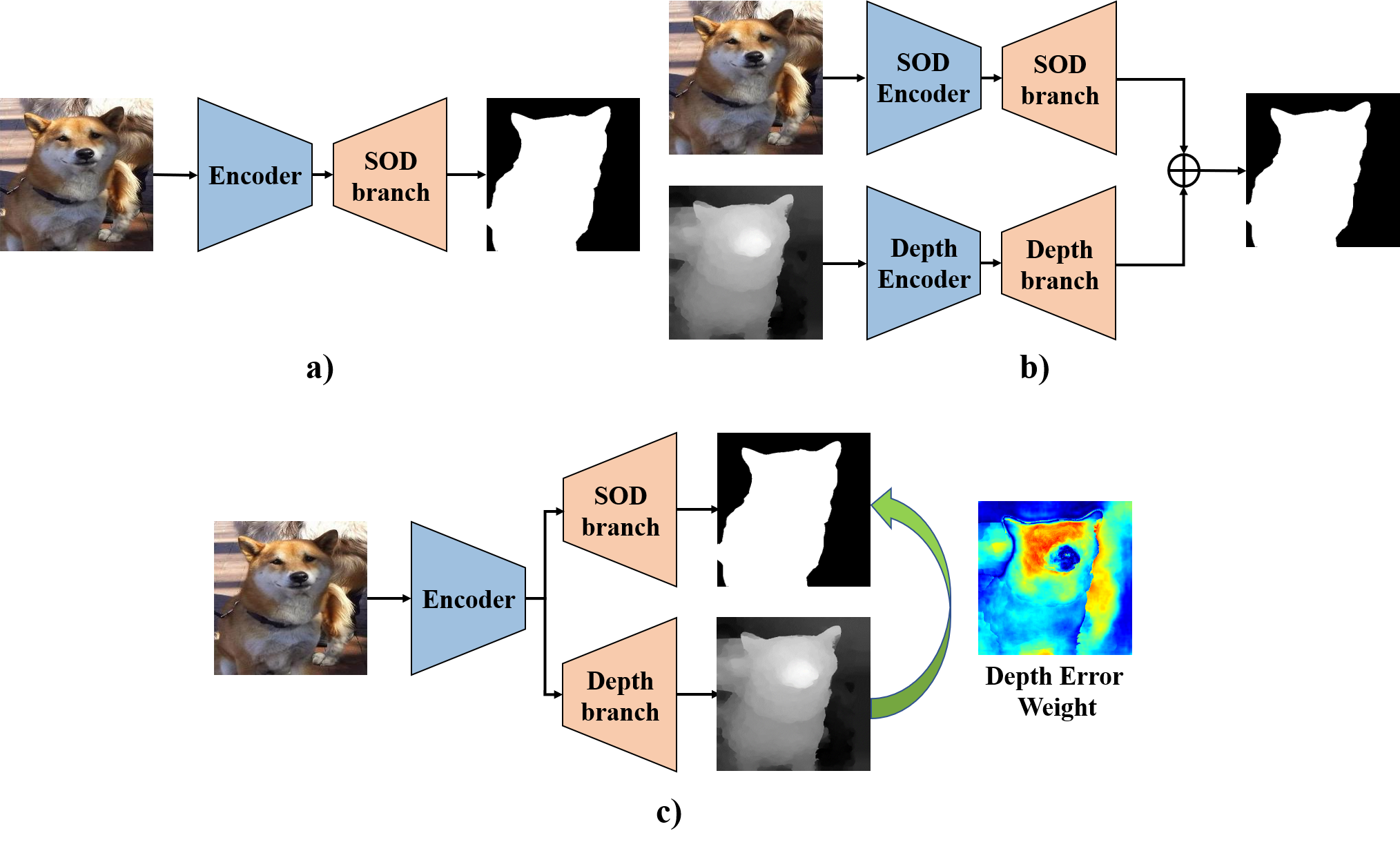}
		\caption{Different types of SOD architecture. a) :  Typical RGB-based SOD network architecture. b): Typical RGBD-based SOD network architecture. c): Proposed Depth-awareness SOD network architecture.
		}\label{fig:introduction}
	\end{center}
\end{figure}

\textbf{Relations and discussions.}
Our intuitive idea comes from the RGB and RGBD salient object detection tasks, which is shown in~\figref{fig:introduction}. The conventional RGB SOD in~\figref{fig:introduction} a) takes the original image as input with a encoder-decoder framework. With the depth as auxiliary input in~\figref{fig:introduction} b), the overall framework requires two independent encoders to extract the depth and RGB features separately, which main computation costs are usually lied on. Moreover, the depth and RGB encoders are separately trained and the relationships between these multi-modal data are not fully explored. 

Taking only RGB inputs as well as saving the computation costs, the depth-aware salient object detection in~\figref{fig:introduction} c) provides us a new perspective to utilize the depth data in this segmentation task. In the testing phase, the network only takes the RGB as input and the object segmentation results are regularized by the depth-awareness constraints in the training phase. In this manner, the network not only builds an explicit relationship between depth and SOD, but also saves the additional costs in feature extraction.

\subsection{Channel-Aware Fusion Module} \label{sec:CAF}
The crucial problem in salient object detection is to select the most discriminative features and pass them in the coarse-to-fine scheme. However, aggregating features from different levels in a encoder-decoder fashion usually leads to missing details or introduces ambiguous features, which both make the network fail to optimize. Notably, this phenomenon appears more frequently when it comes to aggregating features from different domains.
Therefore, a selective feature fusion strategy is in high demand, especially for RGBD salient object understanding.

Toward this end, we propose a novel Channel-Aware Fusion module (CAF), which adaptively selects the discriminative features for object understanding. Instead of using different specific structures for different aggregation strategies in previous works~\cite{su2019selectivity,chen2020global,piao2019depth}, we advocate using a generalized module to fuse any common types of features,~\eg, features from different levels and features from different sources.

\begin{figure}[]
	\begin{center}
		\includegraphics[width=\linewidth]{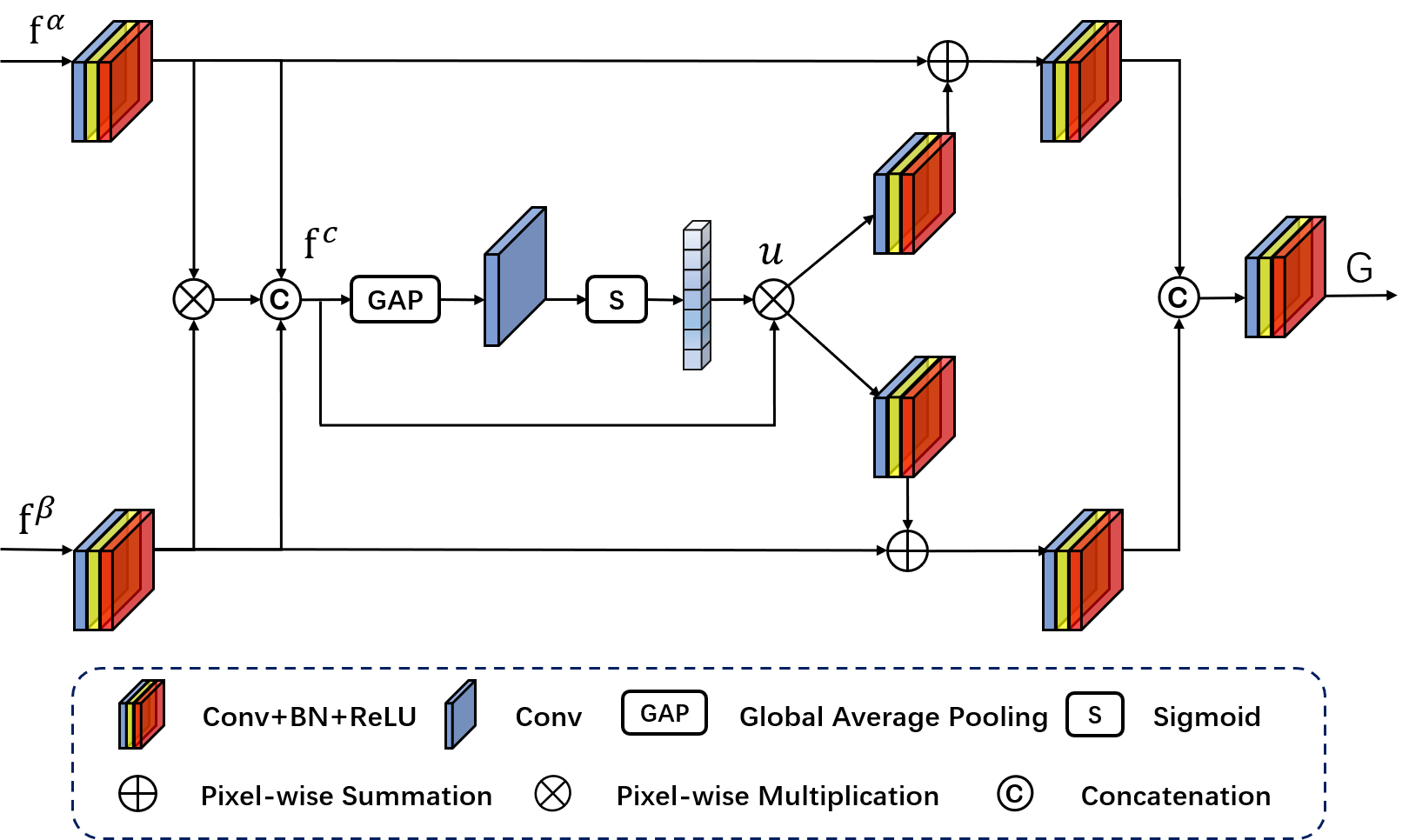}
		\caption{The proposed channel-aware fusion module. Blocks denote basic convolutional units and G is the fused output.
		}\label{fig:CAF}
	\end{center}
\end{figure}

The proposed CAF has some meaningful designs, which are illustrated in~\figref{fig:CAF}. First, given two types of source feature $\br{f}^\alpha, \br{f}^\beta \in \mathbb{R}^{W'\times H'\times C'}$, we use a pixel-wise multiplication to enhance the common pixels in feature maps, while alleviates the ambiguous ones. The enhanced features are then concatenated with the transformed features with a lightweight encoder $\xi(\cdot)$. It can be formally represented as:
\begin{equation} \label{eq:eq feature fusion process1}
\br{f}^{c} = \xi_\alpha(\br{f}^\alpha) \copyright \xi_\beta(\br{f}^\beta) \copyright (\xi_\alpha(\br{f}^\alpha) \otimes \xi_\beta(\br{f}^\beta)),
\end{equation} 
where $\copyright$ and $\otimes$ denote the feature concatenation operation and pixel-wise multiplication respectively. Each encoder $\xi_{\{\alpha,\beta\}}$ is typically composed of a $3\times 3$ convolutional layer followed by a Batch Normalization and a ReLU activation. 
Specially, when aggregating the multi-level features, the features $\br{f}^\alpha$ and $\br{f}^\beta$ are first upsampled to the same scale, which is omitted for better view in~\figref{fig:CAF}. 

After obtaining rich feature $\br{f}^{c}\in \mathbb{R}^{W'\times H'\times 3C'}$ by~\equref{eq:eq feature fusion process1}, the second main concern is how to select the most relevant features that are highly-responded in the segmentation target. Inspired by channel-attention mechanism~\cite{hu2018squeeze,chen2017sca}, we thus propose to use global features for a contextual understanding for the attention weights. The $\br{f}^{c}$ are then squeezed with a global average pooling, followed by a sigmoid normalization $\sigma$, and transformed as the vector shape to align the dimensions with feature channels.
This serialized operation has the form:
\begin{equation}\label{eq:eq feature fusion process2}
\br{g}=\frac{1}{W'\times H'} \sum_{i=1}^{W'}\sum_{j=1}^{H'}\br{f}^{c}_{i,j},
\end{equation}
\begin{equation} \label{eq:eq feature fusion process3}
\br{u}_{i,j} = \br{f}^{c}_{i,j} \otimes \sigma(\varphi_{c}(\br{g}_{i,j})).
\end{equation}
The $\varphi$ is a linear transformation to reorganize the pooling features and $\br{u}$ denotes the learnt attention weighted features. Therefore features relevant to the salient target could be prominent in each group of source features $\br{f}^\alpha$ and $\br{f}^\beta$. This can be achieved by a channel-aware attention mechanism:
\begin{equation} \label{eq:eq feature fusion process3}
\mc{G} = \tau_g (\tau_{v1}(\xi_{\alpha}(f^\alpha) \oplus \xi_{u1}(\br{u}))
\copyright 
\tau_{v2}(\xi_{\beta}(f^\beta) \oplus \xi_{u2}(\br{u}))),
\end{equation}
where $\xi_{\{u1,u2\}}$ denotes the typical decoder and $\tau_{\{v1,v2,g\}}$ denotes the typical decoder with dimensional reductions as original input. Hence the relevant features to target object can be enhanced in the final output $\mc{G}$. 
In addition, to implement the whole framework in a lightweight trend, the channel dimension $C'$ is empirically set as $64$ to achieve the state-of-the-art performance.

\subsection{Depth-awareness Constraint}

What roles does depth play in salient object detection? To answer this aforementioned question, in this paper, we propose an innovative depth-awareness constraint from two complementary aspects,~\ie, multi-level depth awareness and depth error-weighted correction. These two aspects work collaboratively to regularize the salient features being aware of contrastive depth regions and contextual salient confusions, which facilitates the segmentation process in different learning stages.

\textbf{Multi-level depth awareness.} As mentioned in~\secref{sec:CAF}, the key issue in salient object detection lies on the utilization of multi-level features in different network stages. Besides the aggregation strategy, the other exploitation is to regularize the features focusing on meaningful regions, which would provide useful contextual information before aggregation. Taking the advantages of depth information and the hierarchical network architecture, we force the segmentation features to focus on depth regions in different network learning stages, which is elaborated in~\figref{fig:architecture}. This means in each network learning stages, the features should be aware of the object information as well as the contrastive depth regions. We use an additional depth branch to regress the ground-truth depth.

With this collaborative learning of SOD and depth regression, we further fuse these two modules to refine the salient object (see~\figref{fig:architecture}), which builds strong correlations between these two different types of features. Notably, this refinement strategy can also be well handled by our proposed CAF, with the same segmentation supervision at multiple levels. As a result, the salient features stand as a predominant place in the final optimization and the depth map becomes a leading guidance.  
 
\textbf{Depth error-weighted correction.}
To make a thorough exploitation of depth information, we further propose a depth error-weighted correction (DEC) which aims to regularize hard pixels with higher weights if the predicted depth make mistakes. As it stands, the network itself naturally tends to be highly responded to the salient regions and then form a holistic salient object. However, this would guide the predicted depth features focusing on salient regions. This would cause a severe misalignment between the predicted depth and ground truth data. Remarkably, the error regions where the predicted depth make mistakes are usually the semantic ambiguous regions, which we need to pay more attention to the learning process.

\begin{figure}[H]
 	\begin{center}
 		\includegraphics[width= \linewidth]{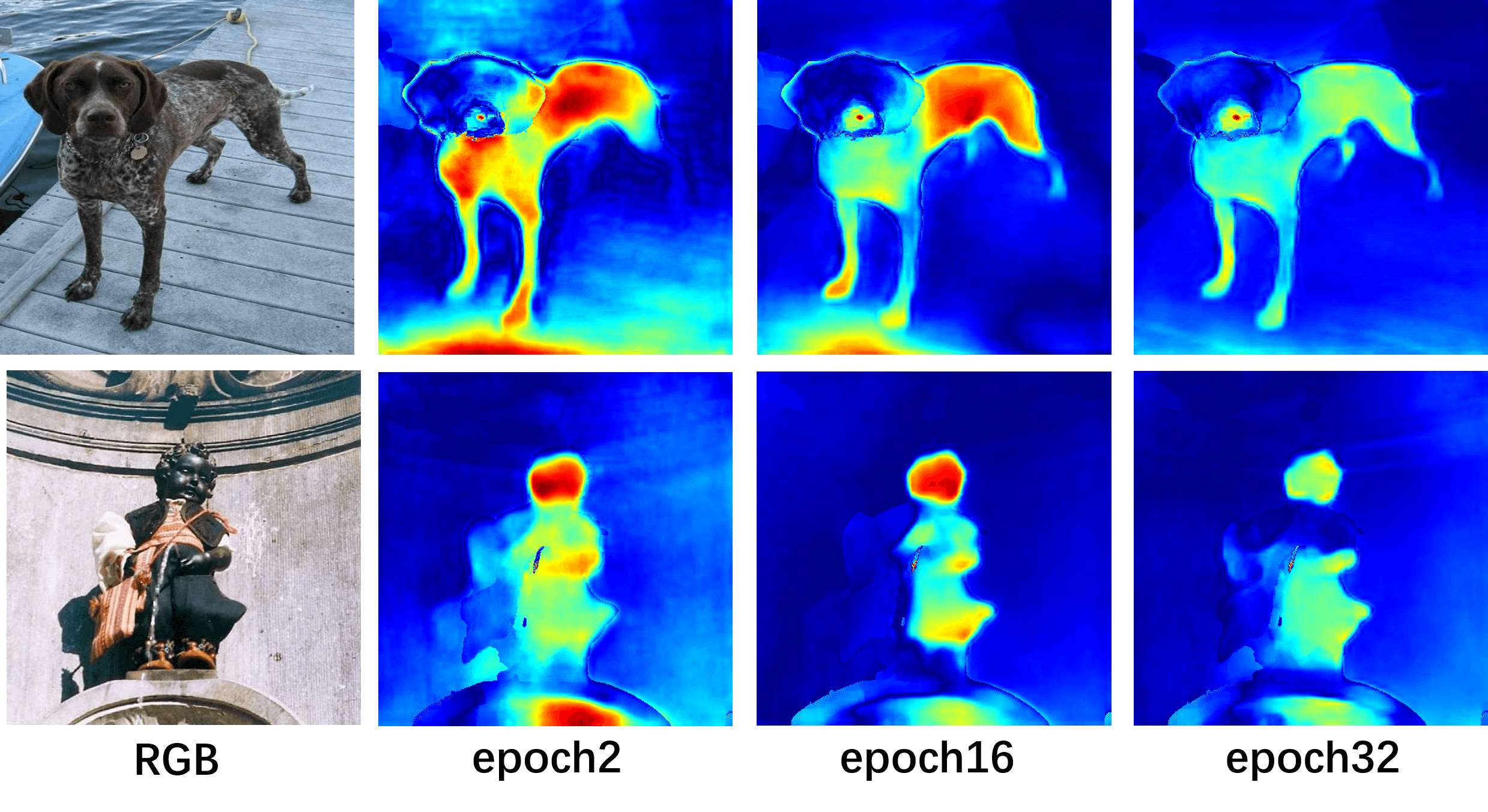}
 		\caption{Qualitative visualization of depth error weights during the training stage, with epoch 2, 16, and 32.
 		}\label{fig:weights}
 	\end{center}
 	
 \end{figure}

In order to solve this misalignment as well as to exploit it, we thus introduce a logarithmic depth error weight. Let $p^d$ and $y^d$ be the predicted depth and groundtruth depth respectively, the error weight $\br{e}_{ij}$ of each pixel has the form:
\begin{equation} \label{eq:dew}
\br{e}_{ij}=\frac{\sum_{i=1}^{h}\sum_{j=1}^{w}(\log p_{ij}^d-\log y_{ij}^d)}{\sum_{i=1}^{h}\sum_{j=1}^{w} \max(\log p^d-\log y^d)},
\end{equation}
where $w$ and $h$ are the width and height of the error window, which aims to represent the error of central pixel with the mean value of a local region. The detailed ablations to decide $w$ and $h$ can be found in~\tabref{table:kernal size}.
In this way, the ambiguous pixels are treated with more attention in the early learning phase. With the optimization goes through, the regularized features become depth-aware and errors are progressively corrected. This learning progress is exhibited in~\figref{fig:weights}, where the highly-responded regions in the error map shrink along with the learning stage. This verifies that the final optimized features are aware of depth information and better at handling semantic confusions.

\subsection{Learning Objective}

Our overall learning objective is composed of three modules, as in~\figref{fig:architecture}, the SOD module, the depth awareness module and the error-weighted correction. Let $p^s, y^s \in \mathbb{R}^{W\times H\times 1}$ be the predicted salient mask and corresponding groudtruth, the SOD module is supervised with the BCE loss:
\begin{equation}\label{eq:bce}
\begin{split}
\mc{L}_{bce} = -\sum_{i=1}^{H}\sum_{j=1}^{W}y^s_{ij}\log(p^s_{ij}).
\end{split}
\end{equation}
However, the BCE loss usually leads to noisy predictions which does not form a holistic object. To make the salient object with clear boundaries, we adopt a IoU (Intersection over Union) loss~\cite{F3Net, qin2019basnet} as the auxiliary loss:
\begin{equation} \label{eq:iou}
\begin{split}
\mc{L}_{iou} = 1-\frac{\sum_{i=1}^{H}\sum_{j=1}^{W} (y^s_{ij}\times p^s_{ij})+1}{\sum_{i=1}^{H}\sum_{j=1}^{W}(y^s_{ij}+p^s_{ij}-y^s_{ij}\times p^s_{ij})+1}.
\end{split}
\end{equation}

For the depth awareness module, we adopt the log mean square error (logMSE) for supervision \cite{eigen2014depth,eigen2015predicting} to generate smooth depth map, and meanwhile providing the error weights $\br{e}$:
\begin{equation} \label{eq:logmse}
\begin{split}
\mc{L}_{depth} = \frac{1}{W\times H}\sum_{i=1}^{H}\sum_{j=1}^{W} || \log y^d_{ij} - \log p^d_{ij} ||^2_2.
\end{split}
\end{equation}

For the error-weighted correction module, we adopt a error-weighted BCE loss to attach more importance to wrongly-predicted pixels:
\begin{equation} \label{eq:wbce}
\mc{L}_{dec} = \frac{-\sum_{i=1}^{H}\sum_{j=1}^{W}\br{e}_{ij}\times y^s_{ij}\log(p^s_{ij})}{\sum_{i=1}^{H}\sum_{j=1}^{W}\br{e}_{ij}}.
\end{equation}
This error loss $\mc{L}_{dec}$ adopts the same supervision as the SOD module with a binary segmentation mask. To implement the multi-level supervision in a unified framework, the overall loss can be formulated as:
\begin{equation} \label{eq:loss}
\begin{split}
\mc{L} = \mc{L}_{depth}+\sum_{i=1}^{S}\lambda_i(\mc{L}_{bce}+\mc{L}_{iou}+\mc{L}_{dec}),
\end{split}
\end{equation}
where $\lambda_i$ denotes the weight of different level loss and $S$ is set as 5 with five stages in ResNet. Here we follow GCPANet~\cite{chen2020global} and set $\lambda$ as [1, 0.8, 0.6, 0.4, 0.2].

\begin{table*}[t]
    \centering{
	\caption{Performance comparison with 9 state-of-the-art RGBD-based SOD methods on five benchmarks. Smaller $MAE$, larger $F_\beta^{max}$, $F_\beta^{mean}$ and $S_\alpha$ indicates better performance. The best results are highlighted in bold.}
	\label{table: RGBD benchmark}
	\resizebox{\textwidth}{!}{
	\begin{tabular}{l|cccc|cccc|cccc|cccc|cccc}
		\hline
		\multicolumn{1}{c|}{\multirow{2}{*}{methods}} &
		\multicolumn{4}{c|}{NJUD-TE} &
		\multicolumn{4}{c|}{NLPR-TE} &
		\multicolumn{4}{c|}{STEREO} &
		\multicolumn{4}{c|}{DES} &
		\multicolumn{4}{c}{SSD} \\ 
		\multicolumn{1}{c|}{} & $F_\beta^{max}$  & $F_\beta^{mean}$    & $MAE$  & $S_\alpha$    & $F_\beta^{max}$   &  $F_\beta^{mean}$    & $MAE$  & $S_\alpha$    & $F_\beta^{max}$   &  $F_\beta^{mean}$    & $MAE$  & $S_\alpha$   & $F_\beta^{max}$   &  $F_\beta^{mean}$    & $MAE$  & $S_\alpha$    & $F_\beta^{max}$  &  $F_\beta^{mean}$    & $MAE$  & $S_\alpha$    \\ \hline
		DF~\cite{qu2017rgbd}                    & .804 & .744 & .141 & .763 & .778 & .682 & .085 & .802 & .757 & .616 & .141 & .757 & .766 & .566 & .093 & .752 & .735 & .709 & .142 & .747 \\
		AFNet~\cite{wang2019adaptive}                 & .775 & .764 & .100 & .772 & .771 & .755 & .058 & .799 & .823 & .806 & .075 & .825 & .728 & .713 & .068 & .770 & .687 & .672 & .118 & .714 \\
		CTMF~\cite{han2017cnns}                & .845 & .788 & .085 & .849 & .825 & .723 & .056 & .860 & .831 & .786 & .086 & .848 & .844 & .765 & .055 & .863 & .729 & .709 & .099 & .776 \\
		MMCI~\cite{chen2019multi}                & .852 & .813 & .079 & .858 & .815 & .729 & .059 & .856 & .863 & .812 & .068 & .873 & .822 & .750 & .065 & .848 & .781 & .748 & .082 & .813 \\
		PCF~\cite{chen2018progressively}                 & .872 & .844 & .059 & .877 & .841 & .794 & .044 & .874 & .860 & .845 & .064 & .875 & .804 & .763 & .049 & .842 & .807 & .786 & .062 & .841 \\
		TANet~\cite{chen2019three}               & .874 & .844 & .060 & .878 & .863 & .796 & .041 & .886 & .861 & .828 & .060 & .871 & .827 & .795 & .046 & .858 & .810 & .767 & .063 & .839 \\
		CPFP\cite{zhao2019contrast}                & .876 & .850 & .053 & .879 & .869 & .840 & .036 & .888 & .874 & .842 & .051 & .879 & .838 & .815 & .038 & .872 & .766 & .747 & .082 & .807 \\
		DMRA~\cite{piao2019depth}                & .886 & .872 & .051 & .886 & .879 & .855 & .031 & .899 & .868 & .847 & .066 & .835 & .888 & .857 & .030 & .900 & .844 & .821 & .058 & .857 \\
		D3Net~\cite{fan2019D3Net}                 & .889 & .860 & .051 & .895 & .885 & .853 & .030 & .904 & .881 & .844 & .054 & .904 & .885 & .859 & .030 & .904 & .847 & .818 & .058 & .866 \\ \hline
		Ours &
		\textbf{.911} &
		\textbf{.894} &
		\textbf{.042} &
		\textbf{.902} &
		\textbf{.929} &
		\textbf{.907} &
		\textbf{.021} &
		\textbf{.929} &
		\textbf{.915} &
		\textbf{.894} &
		\textbf{.037} &
		\textbf{.910} &
		\textbf{.928} &
		\textbf{.892} &
		\textbf{.023} &
		\textbf{.908} &
		\textbf{.881} &
		\textbf{.857} &
		\textbf{.042} &
		\textbf{.885} \\ \hline
	\end{tabular}}
	}
\end{table*}

\begin{figure*}[h]
	\begin{center}
		\includegraphics[width=\textwidth]{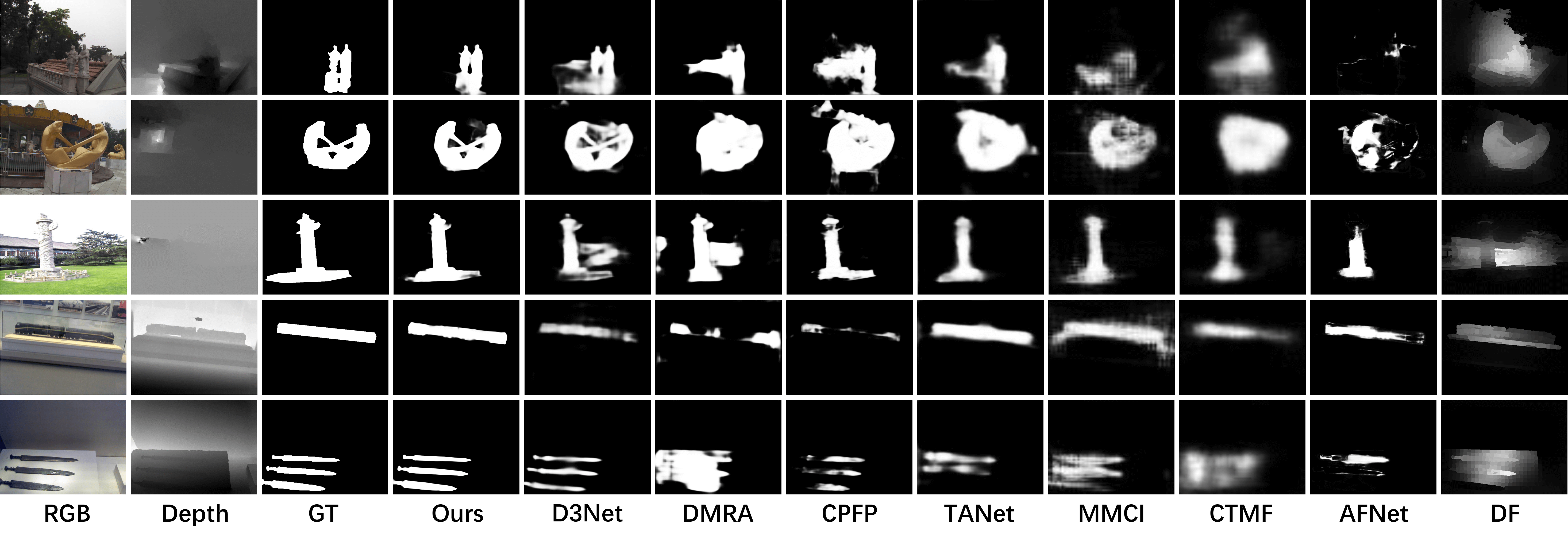}
		\caption{Qualitative comparison of the state-of-the-art RGBD-based methods and our approach. Obviously, saliency maps produced by our model are clearer and more accurate than others in various challenging scenarios.
		}\label{fig:RGBD comparison}
	\end{center}
\end{figure*}

\section{EXPERIMENTS}

\subsection{Datasets and Evaluation Metrics}
\textbf{RGBD-based SOD datasets.} To evaluate the RGBD performance of the proposed approach, we conduct experiments on five benchmarks~\cite{ju2014depth,peng2014rgbd,niu2012leveraging,cheng2014depth,zhu2017three}, including NJUD~\cite{ju2014depth} with 1,985 images captured by Fuji W3 stereo camera, NLPR~\cite{peng2014rgbd} with 1,000 images captured by Kinect. STEREO~\cite{niu2012leveraging} with 1,000 images collected in the Internet.  DES~\cite{cheng2014depth} with 135 images captured by Kinect. SSD~\cite{zhu2017three} with 80 images picked up from stereo movies. Following~\cite{zhao2019contrast,piao2019depth}, We split 1,500 samples from NJUD and 700 samples from NLPR for training, the rest images in these two datasets and the other three datasets are used for testing.

\textbf{RGB-based SOD datasets.} To verify the effectiveness for RGB datasets, we adopt five RGB benchmarks~\cite{wang2017learning,yang2013saliency,yan2013hierarchical,li2014secrets,li2015visual}, including DUTS~\cite{wang2017learning} with 15,572 images, ECSSD~\cite{yan2013hierarchical} with 1,000 images, DUT-OMRON~\cite{yang2013saliency} with 5,168 images, PASCAL-S~\cite{li2014secrets} with 850 images, HKU-IS~\cite{li2015visual} with 4,447 images. DUTS is currently the largest SOD dataset, following~\cite{wang2017learning}, we split 10,553 images (DUT-TR) from DUTS for training and 5,019 images (DUT-TE) from DUTS for testing, the other four datasets are also used for testing.

\textbf{Evaluation Metrics.} To quantitatively evaluate the performance of our approach and state-of-the-art methods, we adopt 4 commonly used metrics: max F-measure ($F_\beta^{max}$), mean F-measure ($F_\beta^{mean}$), mean absolute error ($MAE$) and structure similarity measure ($S_\alpha$)~\cite{fan2017structure} on both RGB-based methods and RGBD-based methods.

We use $F_\beta$ to measure both Precision and Recall comprehensively. $F_\beta$ is computed based on Precision and Recall pairs as follows:
 \begin{equation} \label{eq:F_beta}
 \begin{split}
 F_\beta = \frac{(1+\beta^2)\times Precision\times Recall}{\beta^2\times Precision + Recall},
 \end{split}
 \end{equation}
 where we set $\beta^2$=0.3 to emphasize more on Precision than Recall, and compute $F_\beta^{max}$, $F_\beta^{mean}$ using different thresholds as in \cite{borji2015salient}.
 
 We use $S_\alpha$ to measure structure similarity for a more comprehensive evaluation. $S_\alpha$ combines the region-aware ($S_r$) and object-aware ($S_o$) structural similarity as follows:
  \begin{equation} \label{eq:S_alpha}
 \begin{split}
 S_\alpha = \alpha \times S_o + (1-\alpha)\times S_r,
 \end{split}
 \end{equation}
 where we set $\alpha$=0.5 as suggested in \cite{fan2017structure}.
 
\subsection{Implementation Details}

We adopt ResNet-50~\cite{he2016deep} pre-trained on ImageNet~\cite{deng2009imagenet} as our backbone. The atrous rate of ASPP follows the prior work~\cite{chen2017deeplab}, which is set as (6, 12, 18). In the training stage, we resize each image to $352\times 352$ and adopt horizontal flip, random crop and multi-scale resize as data augmentation. We use SGD optimizer with the batch size=32 for 32 epochs. Inspired by~\cite{F3Net,chen2020global}, we adopt warm-up and linear decay strategies to adjust the learning rate with the maximum learning rate 0.005 for ResNet-50 backbone and 0.05 for other parts. We set momentum and decay rate to 0.9 and 5e-4, respectively. It only takes us 1 hour for RGBD-based task and 3 hours for RGB-based task to train a model on one NVIDIA 1080Ti GPU. 

For the RGBD-based salient object detection, we utilize both RGB images and depth maps from training sets to train our model. During the testing stage, we only need RGB images as inputs to predict saliency maps on RGBD test sets. 
For the RGB-based salient object detection task, we first estimate depth maps for DUT-TR by pre-trained VNLNet~\cite{yin2019enforcing} directly, which works well in single image depth estimation task. Then we utilize both DUT-TR and its corresponding predicted depth maps to train our model. During the inference stage, we only need RGB images as inputs to predict saliency maps on RGB test sets. The PyTorch implementation will be publicly available.\footnote{Link is masked for blind review policy.}

\begin{table*}[t]
    \centering{
	\caption{Performance comparison with 10 state-of-the-art  RGB-based SOD methods on five benchmarks. Smaller $MAE$, larger $F_\beta^{max}$,$F_\beta^{mean}$ and $S_\alpha$ correspond to better performance. The best results are highlighted in bold.}
	\label{table:RGB benchmark}
	\resizebox{\textwidth}{!}{
		\begin{tabular}{l|cccc|cccc|cccc|cccc|cccc}
			\hline
			\multicolumn{1}{c|}{\multirow{2}{*}{methods}} &
			\multicolumn{4}{c|}{ECSSD} &
			\multicolumn{4}{c|}{DUT-TE} &
			\multicolumn{4}{c|}{DUT-OMRON} &
			\multicolumn{4}{c|}{HKU-IS} &
			\multicolumn{4}{c}{PASCAL-S} \\ 
			\multicolumn{1}{c|}{} &
			$F_\beta^{max}$ &
			$F_\beta^{mean}$ &
			$MAE$ &
			$S_\alpha$ &
			$F_\beta^{max}$ &
			$F_\beta^{mean}$ &
			$MAE$ &
			$S_\alpha$ &
			$F_\beta^{max}$ &
			$F_\beta^{mean}$ &
			$MAE$ &
			$S_\alpha$ &
			$F_\beta^{max}$ &
			$F_\beta^{mean}$ &
			$MAE$ &
			$S_\alpha$ &
			$F_\beta^{max}$ &
			$F_\beta^{mean}$ &
			$MAE$ &
			$S_\alpha$ \\ \hline
			BMPM~\cite{zhang2018bi} &
			.929 &
			.894 &
			.045 &
			.911 &
			.851 &
			.762 &
			.049 &
			.861 &
			.774 &
			.698 &
			.064 &
			.808 &
			.921 &
			.875 &
			.039 &
			.905 &
			.862 &
			.803 &
			.073 &
			.840 \\
			PAGR~\cite{zhang2018progressive} &
			.927 &
			.894 &
			.061 &
			.889 &
			.854 &
			.784 &
			.056 &
			.838 &
			.771 &
			.711 &
			.071 &
			.775 &
			.918 &
			.886 &
			.048 &
			.887 &
			.854 &
			.803 &
			.094 &
			.815 \\
			R3Net~\cite{deng2018r3net} &
			.929 &
			.883 &
			.051 &
			.910 &
			.829 &
			.716 &
			.067 &
			.837 &
			.793 &
			.690 &
			.067 &
			.819 &
			.910 &
			.853 &
			.047 &
			.894 &
			.837 &
			.775 &
			.101 &
			.809 \\
			PiCA-R~\cite{liu2018picanet} &
			.935 &
			.901 &
			.047 &
			.918 &
			.860 &
			.816 &
			.051 &
			.868 &
			.803 &
			.762 &
			.065 &
			.829 &
			.919 &
			.880 &
			.043 &
			.905 &
			.881 &
			.851 &
			.077 &
			.845 \\
			BANet~\cite{su2019selectivity} &
			.939 &
			.917 &
			.041 &
			.924 &
			.872 &
			.829 &
			.040 &
			.879 &
			.782 &
			.750 &
			.061 &
			.832 &
			.923 &
			.893 &
			.037 &
			.913 &
			.847 &
			.839 &
			.079 &
			.852 \\
			PoolNet~\cite{liu2019simple} &
			.944 &
			.915 &
			.039 &
			.921 &
			.880 &
			.809 &
			.040 &
			.883 &
			.808 &
			.747 &
			.055 &
			.833 &
			.933 &
			.899 &
			.032 &
			.916 &
			.869 &
			.822 &
			.074 &
			.845 \\
			BASNet~\cite{qin2019basnet} &
			.943 &
			.880 &
			.037 &
			.916 &
			.859 &
			.791 &
			.048 &
			.866 &
			.805 &
			.756 &
			.056 &
			.836 &
			.928 &
			.895 &
			.032 &
			.909 &
			.857 &
			.775 &
			.078 &
			.832 \\
			CPD-R~\cite{wu2019cascaded} &
			.939 &
			.917 &
			.037 &
			.918 &
			.865 &
			.805 &
			.043 &
			.869 &
			.797 &
			.747 &
			.056 &
			.825 &
			.925 &
			.891 &
			.034 &
			.905 &
			.864 &
			.824 &
			.072 &
			.842 \\
			F3Net~\cite{F3Net} &
			.945 &
			.925 &
			.033 &
			.924 &
			.890 &
			.840 &
			.035 &
			.888 &
			.813 &
			.766 &
			.053 &
			.838 &
			.937 &
			.910 &
			.028 &
			.917 &
			.880 &
			.840 &
			\textbf{.064} &
			.855 \\
			GCPANet~\cite{chen2020global} &
			.948 &
			.919 &
			.035 &
			\textbf{.927} &
			.888 &
			.817 &
			.040 &
			.891 &
			.812 &
			.748 &
			.056 &
			.839 &
			.938 &
			.898 &
			.031 &
			.920 &
			.876 &
			.836 &
			\textbf{.064} &
			\textbf{.861} \\ \hline
			Ours &
			\textbf{.950} &
			\textbf{.932} &
			\textbf{.032} &
			\textbf{.927} &
			\textbf{.896} &
			\textbf{.853} &
			\textbf{.034} &
			\textbf{.894} &
			\textbf{.827} &
			\textbf{.783} &
			\textbf{.050} &
			\textbf{.845} &
			\textbf{.942} &
			\textbf{.917} &
			\textbf{.027} &
			\textbf{.922} &
			\textbf{.885} &
			\textbf{.849} &
			\textbf{.064} &
			.860 \\ \hline
	\end{tabular}}
	}
\end{table*}

\begin{figure*}[h]
	\begin{center}
		\includegraphics[width=1\textwidth]{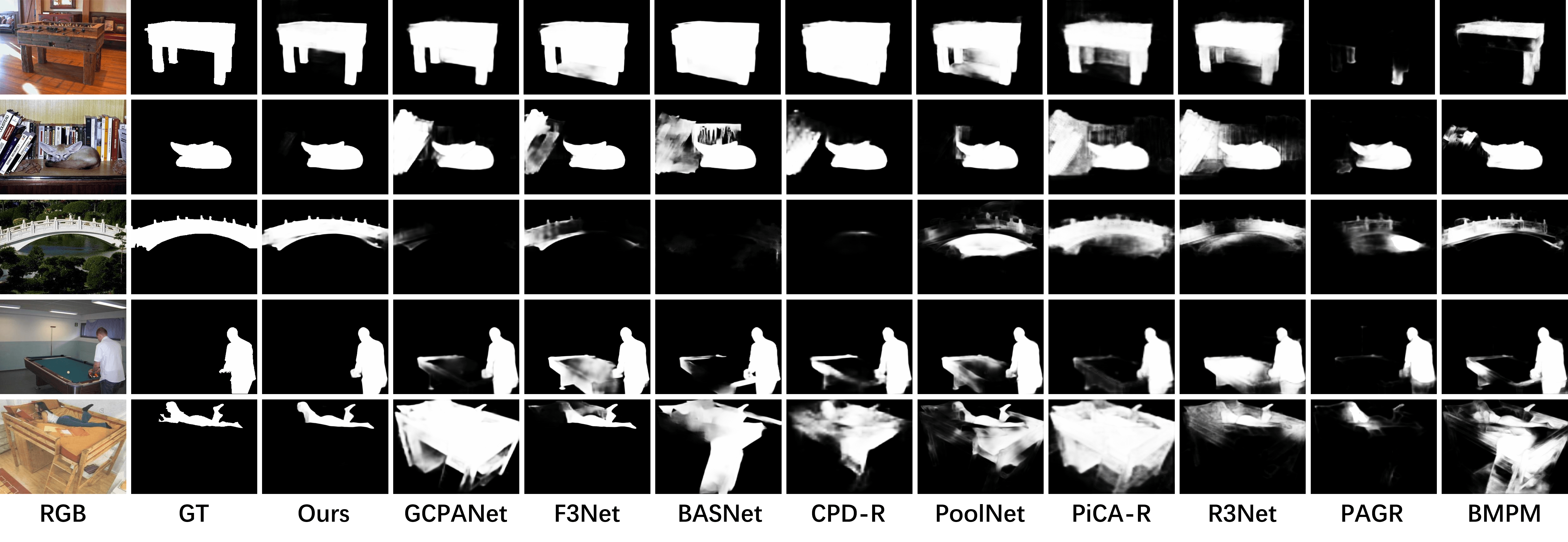}
		\caption{Qualitative comparison of the state-of-the-art RGB-based methods and our approach. Obviously, saliency maps produced by our model are clearer and more accurate than others in various challenging scenarios.
		}\label{fig:RGB comparison}
	\end{center}
\end{figure*}

\subsection{Comparisons with the state-of-the-art}

\textbf{RGBD-based SOD Benchmark.} As shown in \tabref{table: RGBD benchmark}, we compare our model denoted as DASNet with 9 state-of-the art methods, including DF~\cite{qu2017rgbd}, AFNet~\cite{wang2019adaptive}, CTMF~\cite{han2017cnns}, MMCI~\cite{chen2019multi}, PCF~\cite{chen2018progressively}, TANet~\cite{chen2019three}, CPFP~\cite{zhao2019contrast}, DMRA~\cite{piao2019depth}, D3Net~\cite{fan2019D3Net}. For  fair comparisons, we obtain the saliency maps from the reported results. Our proposed approach surpasses 9 state-of-the-art RGBD-based saliency detection methods on five benchmarks. As shown in~\tabref{table: RGBD benchmark}, it is obviously that our method achieves a new performance leader-board with no depth image as inputs, which puts our model in inferior places for comparison.
Especially for the $F_\beta^{max}$ and $F_\beta^{mean}$ metric, our model outperforms over 3\%, which means our method has a good capability to utilize depth information for more precise saliency maps.

In \figref{fig:RGBD comparison}, we exhibit the saliency maps predicted by our model and other approaches. Among all the methods, our model performs best both on completeness and clarity. In the first, second, and third rows, our method could obtain more accurate and clearer saliency maps than others with ambiguous depth cues. In forth and fifth rows, our method could obtain more complete results than others. Our proposed framework could utilize depth cues much better in various challenging scenarios. Besides, the object boundaries predicted by our model are clearer and sharper than others.

\textbf{RGB-based SOD Benchmark.} As shown in \tabref{table:RGB benchmark}, we compare our proposed DASNet with 10 state-of-the-art methods,~\ie, BMPM~\cite{zhang2018bi}, PAGR~\cite{zhang2018progressive}, R3Net~\cite{deng2018r3net}, PiCANet~\cite{liu2018picanet}, PoolNet~\cite{liu2019simple}, BANet~\cite{su2019selectivity}, CPDNet~\cite{wu2019cascaded}, BASNet~\cite{qin2019basnet}, F3Net~\cite{F3Net}, GCPANet~\cite{chen2020global}.
As shown in~\tabref{table:RGB benchmark}, we can see our proposed DSANet still outperforms other methods and ranks first on all datasets and almost all metrics. However, this performance is achieved with only estimated depth maps as training priors. we believe that with the captured real data, the final performance would be improved steadily, which is vailidated on the RGBD benchmarks.

As shown in \figref{fig:RGB comparison}, comparing with visual results of different methods, our approach shows an advantage in completeness and clarity. In first and second rows, our method could distinguish foreground and background and obtain more accurate results than other methods in complex scenarios with similar foreground and background. In third row, our method could obtain more complete results in complex scenarios with low contrast, while other methods might fail to detect salient objects in the same scenarios. In forth and fifth rows, our method can provide accurate object localization when salient objects touching image boundaries. Besides, the object boundaries predicted by our model are clearer and sharper than others.

\begin{table}[t] 
	\caption{Ablation study for different components. BCE, IoU, DEC are different loss functions mentioned above. CAF denotes the proposed channel aware fusion module. DAM denotes the depth awareness module. MLS represents multi-level supervision. }
	\label{table:ablation}
	\begin{tabular}{cccccc|cc}
		\hline
		\multirow{2}{*}{BCE} & \multirow{2}{*}{CAF} & \multirow{2}{*}{IoU} & \multirow{2}{*}{DAM} & \multirow{2}{*}{DEC} & \multirow{2}{*}{MLS} & \multicolumn{2}{c}{NJUD-TE} \\
		&   &   &   &   &   & $F_\beta^{mean}$    & $MAE$  \\ \hline
		\checkmark &   &   &   &   &   & .838 & .058 \\
		\checkmark & \checkmark &   &   &   &   & .853 & .056 \\
		\checkmark & \checkmark &   & \checkmark &   &   & .857 & .051 \\
		\checkmark & \checkmark &   & \checkmark & \checkmark &   & .871 & .048 \\ \hline
		\checkmark & \checkmark & \checkmark &   &   &   & .875 & .047 \\
		\checkmark & \checkmark & \checkmark & \checkmark &   &   & .880 & .045 \\
		\checkmark & \checkmark & \checkmark & \checkmark & \checkmark &   & .886 & .043 \\
		\checkmark & \checkmark & \checkmark & \checkmark & \checkmark & \checkmark & .894 & .042 \\ \hline
	\end{tabular}
\end{table}

\begin{figure}[t]
	\begin{center}
		\includegraphics[width= \linewidth]{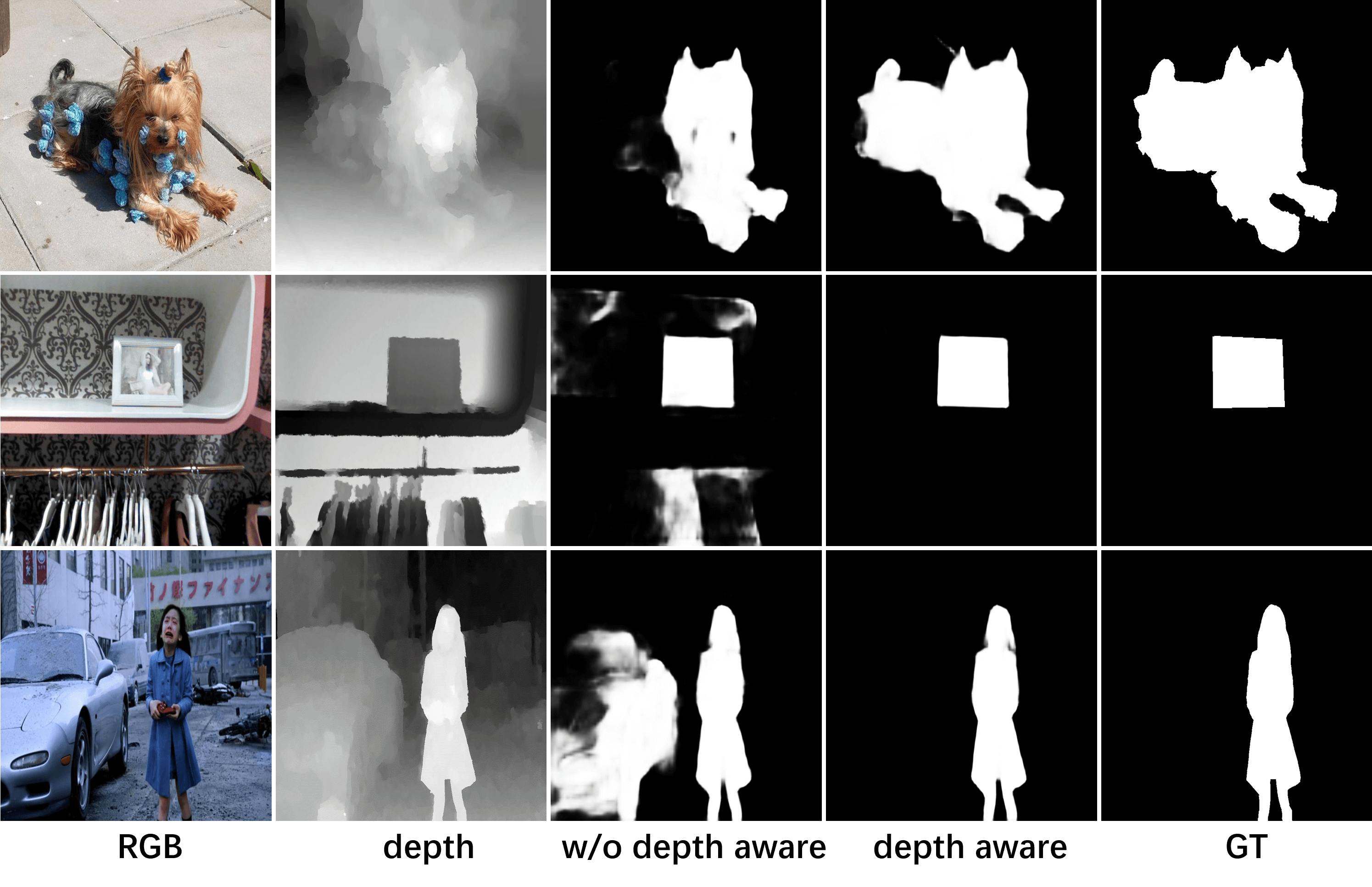}
		\caption{Qualitative results on RGBD datasets. The third column without depth awareness is hard to distinguish complex scenarios with similar foreground and background, while our model in the forth column shows better performance.
		}\label{fig:analysis}
	\end{center}
\end{figure}

\subsection{Performance Analysis}

To investigate the effectiveness of each key component in our proposed model, we first conduct a thorough ablation study and then measure the computation complexity for the state-of-the-art models to show its superiority. Finally an experiment for finding hyper-parameters can be found in~\tabref{table:kernal size}.

\textbf{Channel-Aware Fusion.} To evaluate the effectiveness of our feature fusion module, we reconstruct our model with different ablation factors. \tabref{table:ablation} shows the ablations on NJUD-TE dataset. In the first row, we first build our model with widely-used lateral connections between different levels of features, and then fuse them by pixel-wise summation as our baseline. In the second row, we replace the fusion strategy aforementioned with proposed CAF. This more effective fusion strategy can improve $F_\beta^{mean}$ of baseline from 0.838 to 0.853.

\textbf{Depth-awareness Constraint.} Then we test our proposed DAM and DEC on the baseline using only BCE loss , and both BCE and IoU loss respectively. Comparing with model using CAF and only BCE loss, our proposed DAM and DEC can improve $F_\beta^{mean}$ 1.8\% in total. Compared with the baseline using CAF and both BCE loss and IoU loss, our proposed DAM and DEC can improve $F_\beta^{mean}$ from 0.875 to 0.886 and $MAE$ from 0.047 to 0.043. At last, we add multi-level supervision to refine our results. As shown in \tabref{table:ablation}, all components contribute to the performance improvement, which demonstrates the necessity of each component of our proposed model to obtain the best saliency detection results.
Qualitative results can be found in \figref{fig:analysis}. In the third column, our model without the DAM and DEC would be confused in regions with similar foreground and background. With DAM and DEC, our model could distinguish these confusing features and generate more accurate and clearer saliency maps.

\textbf{Computational efficiency.} \tabref{table:params} shows the parameters and computational cost measured by multiply-adds (MAdds) of our proposed model and other open-sourced RGB-based models and RGBD-based models. Our model could achieve obvious higher performance in a light-weight fashion. Notably, CPD-R~\cite{wu2019cascaded} discards features of two shallower layers to improve the computation efficiency, but sacrifices the accuracy and clarity of results. For fair comparisons, we obtain the deployment codes released by authors and evaluate them with the same configuration.

\textbf{Hyper-paramters.} To evaluate the effectiveness as well as to find the adequate window sizes in~\equref{eq:dew}, we tune the $w\times h$ to be different sizes and choose $7\times7$ to achieve the best performance. This means that the error weight should be locally aware thus to generate clear object details.  
This also indicates that amplifying the local receptive field of error-weighted correction module in an adequate range is effective to reach higher scores.

\begin{table}[t]
\setlength{\tabcolsep}{1.5mm}
	\caption{Complexity comparison with RGB-based models and RGBD-based models. Models ranking the first and second place are viewed in bold and underlined.}
	\label{table:params}
	\begin{tabular}{ccccc}
		\hline
		& Methods & Platform & Params(M)      & MAdds(G)       \\ \hline
		RGB\&RGBD              & Ours    & pytorch  & \textbf{36.68} & {\underline{11.57}}   \\ \hline
		\multirow{4}{*}{RGB}  & GCPANet~\cite{chen2020global} & pytorch  & 67.06          & 26.61         \\
		& BASNet~\cite{qin2019basnet}  & pytorch  & 87.06          & 97.51         \\
		& CPD-R~\cite{wu2019cascaded}   & pytorch  & {\underline{47.85}}    & \textbf{7.19} \\
		& BANet~\cite{su2019selectivity}   & caffe    & 55.90          & 35.83         \\ \hline
		\multirow{2}{*}{RGBD} & CPFP~\cite{zhao2019contrast}    & caffe    & 72.94          & 21.25         \\
		& DMRA~\cite{piao2019depth}    & pytorch  & 59.66          & 113.09        \\ \hline
	\end{tabular}
\end{table}

\begin{table}[t]
\setlength{\tabcolsep}{2.8mm}
	\caption{Error correction results on NLPR-TE with different window sizes.}
	\label{table:kernal size}
	\begin{tabular}{c|cccccc}
		\hline
		& $1\times1$ & $3\times3$ & $5\times5$ & $7\times7$ & $15\times15$ & $31\times31$  \\ \hline
		$F_\beta^{max}$ &  .924   &  \textbf{.929}   &  .925   &  \textbf{.929}   &  .926  &  .927         \\ 
		$F_\beta^{mean}$  &  .895   &  .904   &  .898   &  \textbf{.907}   &  .904 &   .897          \\ 
		$MAE$  &  .024   &  \textbf{.021}   &  .022   &  \textbf{.021}   &   .023  & .022  \\
		$S_\alpha$ &  .924   &  .928   &  .926   &  \textbf{.929}   &   .926 & .925        \\ \hline
	\end{tabular}
\end{table}

\section{CONCLUSIONS}
In this paper, we rethink the problem of depth in the field of salient object detection and propose a new perspective of containing the depth constraints in learning process, rather than using the captured depth as inputs. 
To make a deeper exploitation of depth information, we develop a multi-level depth awareness constraints and a depth error-weighted loss to alleviate the salient confusions. These advanced designs endow our model lightweight and be free of depth input.
Experimental results reveal that with only RGB inputs, the proposed network not only surpasses the state-of-the-art RGBD methods by a large margin but well demonstrates its effectiveness in RGB application scenarios.

\bibliographystyle{ACM-Reference-Format}
\bibliography{main}

\appendix

\end{document}